\documentclass[11pt,a4paper]{article}
\PassOptionsToPackage{breaklinks}{hyperref}

\usepackage[hyperref]{emnlp-ijcnlp-2019}
\usepackage{times}
\usepackage{latexsym}

\usepackage{multirow}
\usepackage{graphicx}
\usepackage{subcaption}
\usepackage{url}
\usepackage{algorithm}
\usepackage{algpseudocode}

\aclfinalcopy

\newcommand{\Sec}[1]{{Section~\ref{section:#1}}}
\newcommand{\Fig}[1]{{Figure~\ref{figure:#1}}}
\newcommand{\Tab}[1]{{Table~\ref{table:#1}}}
\newcommand{\Algo}[1]{{Algorithm~\ref{algo:#1}}}
\newcommand{\NxM}{{$N\times M$}}
\newcommand{\task}[2]{{#1$\rightarrow$#2}}

\title{Multi-Layer Softmaxing during Training Neural Machine Translation\\ for Flexible Decoding with Fewer Layers}

\author{Raj Dabre \qquad Atsushi Fujita\\
        National Institute of Information and Communications Technology \\
        3-5 Hikaridai, Seika-cho, Soraku-gun, Kyoto, 619-0289, Japan\\
        \textsf{firstname.lastname@nict.go.jp}
}

\date{}

\begin{document}
\maketitle
\begin{abstract}
This paper proposes a novel procedure for training an encoder-decoder based deep neural network which compresses $N\times M$ models into a single model enabling us to dynamically choose the number of encoder and decoder layers for decoding. Usually, the output of the last layer of the $N$-layer encoder is fed to the $M$-layer decoder, and the output of the last decoder layer is used to compute softmax loss. Instead, our method computes a single loss consisting of $N\times M$ losses: the softmax loss for the output of each of the $M$ decoder layers derived using the output of each of the $N$ encoder layers. A single model trained by our method can be used for decoding with an arbitrary fewer number of encoder and decoder layers. In practical scenarios, this (a)~enables faster decoding with insignificant losses in translation quality and (b)~alleviates the need to train $N\times M$ models, thereby saving space. We take a case study of neural machine translation and show the advantage and give a cost-benefit analysis of our approach.
\end{abstract}

\section{Introduction}
\label{section:introduction}

Deep neural networks, which allow for end-to-end training, typically consist of an encoder and a decoder coupled via an attention mechanism. Whereas the very first deep models used stacked recurrent neural networks (RNN) \cite{DBLP:journals/corr/SutskeverVL14,DBLP:journals/corr/ChoMGBSB14,DBLP:journals/corr/BahdanauCB14:original} in the encoder and decoder, the recent Transformer model \cite{NIPS2017_7181} constitutes the current state-of-the-art approach, owing to its better context generation mechanism via multi-head self- and cross-attentions.

Given an encoder-decoder architecture and its hyper-parameters, such as the number of layers of encoder and decoder and the sizes of vocabularies (in the case of text based models) and hidden layers, the parameters of the model, i.e., matrices and biases for non-linear transformations, are optimized by iteratively updating them so that the loss for the training data is minimized.  The hyper-parameters can also be tuned, for instance, through maximizing the automatic evaluation score on the development data.
However, in general, it is not guaranteed (and also highly impossible) that a single set of hyper-parameters suffices diverse cost-benefit demands at the same time.  For instance, in practical low-latency scenarios, it is often acceptable to sacrifice output quality for speed.  Once a model has been trained, using fewer number of layers for faster decoding is theoretically possible.
Note also that an optimal set of hyper-parameters does not guarantee that it always results in the best translation for any input.  Hosting multiple models simultaneously for flexible decoding is impractical, since it requires unreasonably large quantity of memory.

\begin{figure*}[t]
    \centering
    \begin{subfigure}[t]{0.48\textwidth}
    \centering
    \includegraphics[scale=.4]{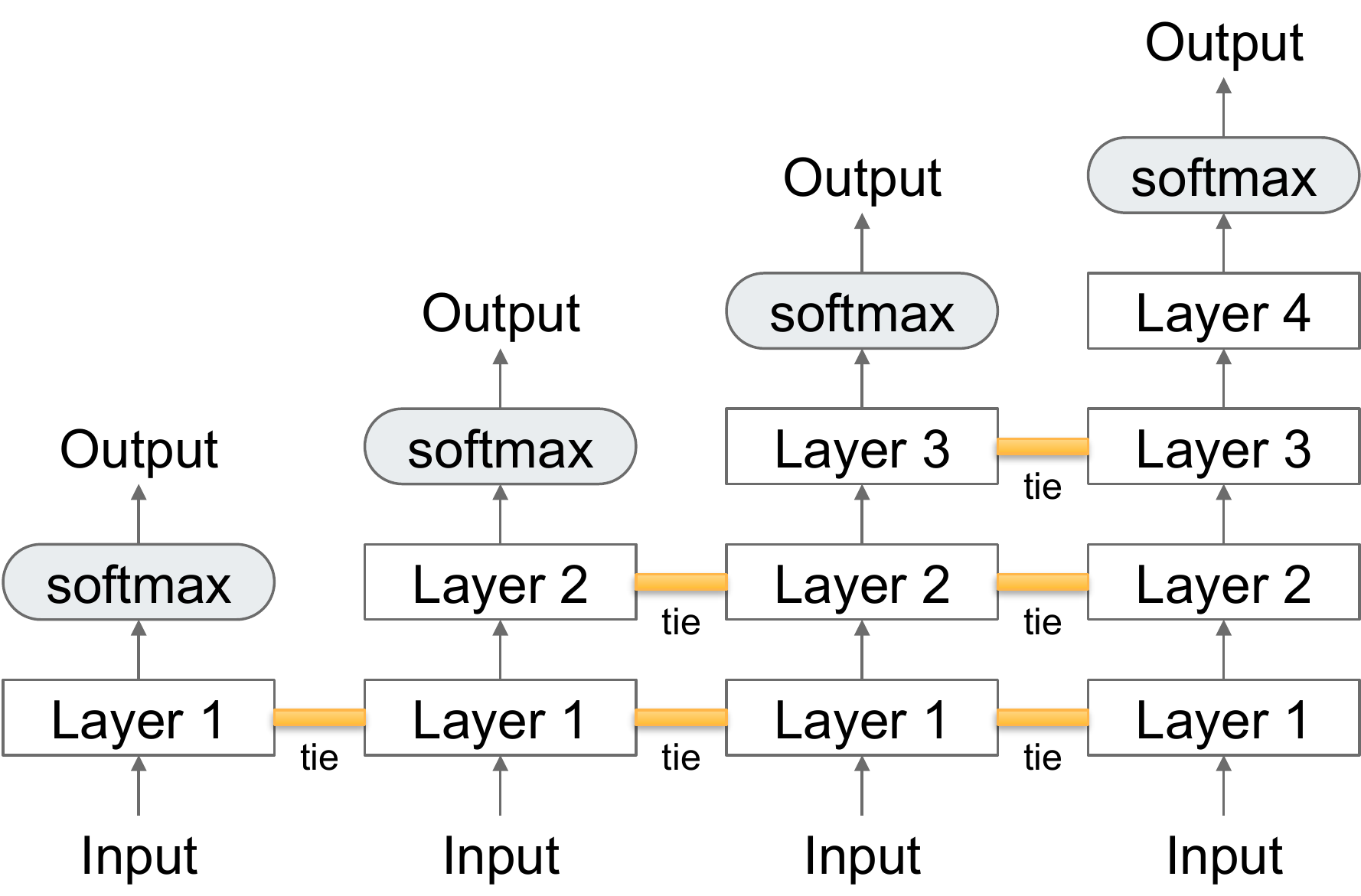}
    \caption{Multiple tied-layer vanilla models.}
    \label{figure:tie4}
    \end{subfigure}
    \begin{subfigure}[t]{0.48\textwidth}
    \centering
    \includegraphics[scale=.4]{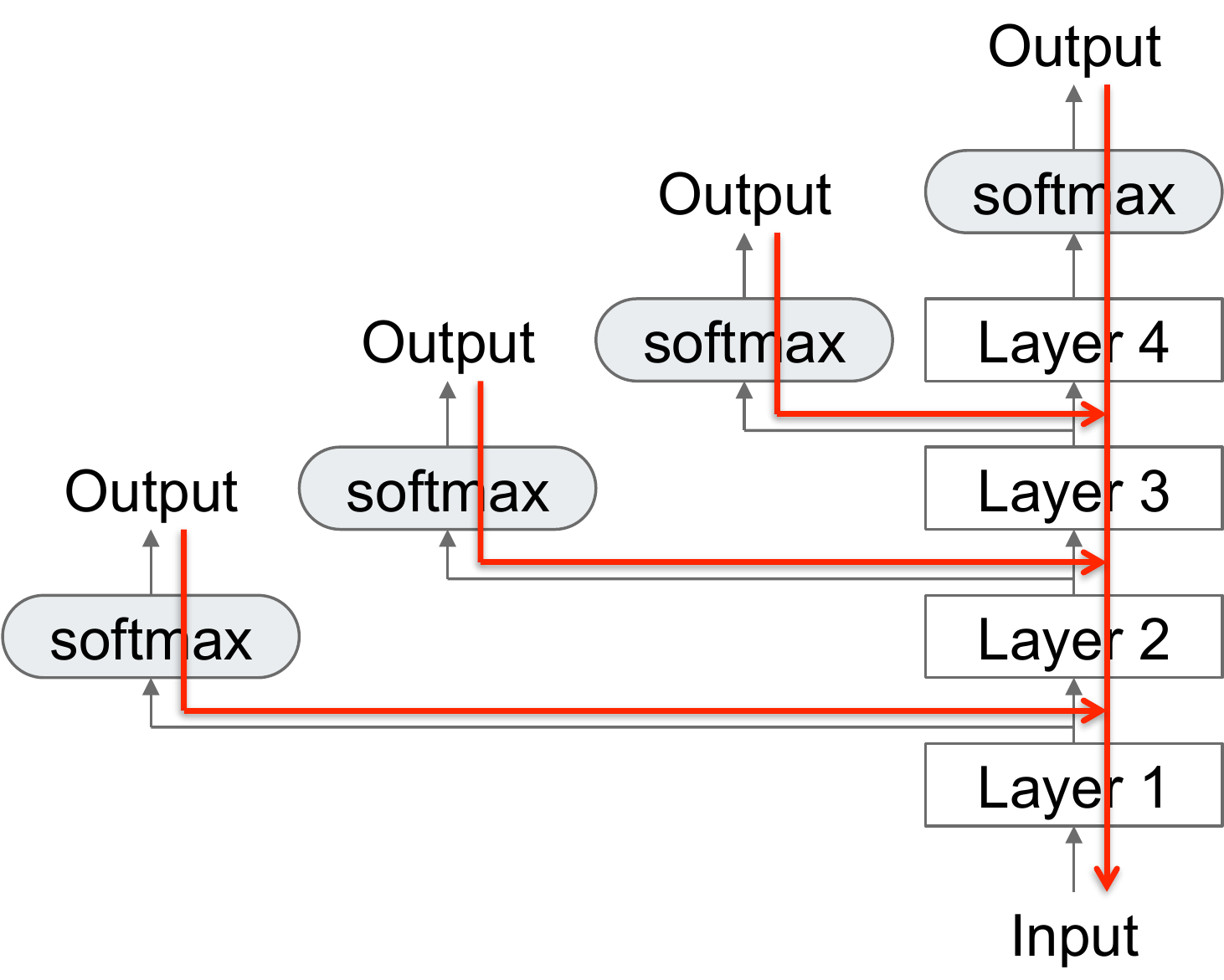}
    \caption{Collapsing tied layers into one.}
    \label{figure:tie4simple}
    \end{subfigure}
    \caption{The general concept of multi-layer softmaxing for training multi-layer neural models with an example of a 4-layer model. \Fig{tie4} is a depiction of our idea in the form of multiple vanilla models whose layers are tied together.  \Fig{tie4simple} shows the result of collapsing all tied layers into a single layer.  The red lines indicate the flow of gradients and hence the lowest layer in the stack receives the most updates.}

    \label{figure:multisoftmaxing}
\end{figure*}

To this end, we propose to train multi-layer neural models referring to the output of all layers during training.  Conceptually, this approach equals to tying the parameters of multiple models with different number of layers, as illustrated in \Fig{multisoftmaxing}, and is not specific to any type of multi-layer neural models.
In this paper, however, we specifically focus on encoder-decoder models with $N$ encoder and $M$ decoder layers, and compress {\NxM} models\footnote{Rather than casting the encoder-decoder model into a single column model with ($N+M$) layers.} to update the model, where a total of {\NxM} losses are computed by softmaxing the output of each of the $M$ decoder layers, where it attends to the output of each of the $N$ encoder layers. Each decoder layer is updated referring to a direct signal from the overall loss, and so does each encoder layer from all the $M$ decoder layers.  The number of parameters of the resultant encoder-decoder model is equivalent to that of the most complex subsumed model with $N$ encoder and $M$ decoder layers.  Yet, we can now perform faster decoding using a fewer number of encoder and decoder layers, given that shallower layers are better trained.

In this paper, we take the case study of neural machine translation (NMT) \cite{DBLP:journals/corr/ChoMGBSB14,DBLP:journals/corr/BahdanauCB14:original}, where we focus on the numbers of encoder and decoder layers of the Transformer model \cite{NIPS2017_7181}, and demonstrate that it is possible to train a single model with $N$ encoder and $M$ decoder layers that can be used for decoding with flexibly fewer number of layers than $N$ and $M$ without appreciable quality loss.
We evaluate our proposed approach on WMT18 English-to-German translation task, and give a cost-benefit analysis for translation quality vs. decoding speed.

Although we apply our method to encoder-decoder models and evaluate it on an NMT task, the method should potentially be applicable to any general multi-layer neural models.

\section{Related Work}
\label{section:relwork}

There are studies that exploit multiple layers simultaneously.  \citet{multilayer} fused hidden representations of multiple layers in order to improve the translation quality.  \citet{I17-1001} and \citet{eachlayer} focused on identifying which encoder or decoder layer can generate useful representations for different natural language processing tasks.
There are also notable approaches for speeding-up: knowledge distillation \cite{DBLP:journals/corr/HintonVD15,DBLP:journals/corr/FreitagAS17}, average attention networks \cite{DBLP:conf/acl/XiongZS18}, and binary code prediction \cite{P17-1079}.

However, to the best of our knowledge, none of them has tackled the issue in training a flexible translation model.

\section{Multi-Layer Softmaxing}
\Fig{multisoftmaxing} gives a simple overview of the concept of multi-layer softmaxing for training a generic 4-layer model. This model takes an input, passes it through 4 layers,\footnote{We make no assumptions about the nature of the layers.}
and then into the softmax layer to predict the output. Typically, one would apply softmax to the 4th layer only, compute loss, and then back-propagate gradients for updating weights. Instead, we propose to apply softmax to each layer, aggregate the computed losses, and then back-propagate losses. This ensures that during decoding we can choose any layer instead of only the topmost layer.

Extending this to a multi-layer encoder-decoder model is straightforward.  In encoder-decoder models, the encoder comprises an embedding layer for the input (source language for NMT) and $N$ stacked transformation layers. The decoder consists of an embedding layer and a softmax layer for generating the output (target language for NMT) along with $M$ stacked transformation layers.
Let $X$ be the input to the $N$-layer encoder, $Y$ the anticipated output of the $M$-layer decoder as well as the input to the decoder (for training), and $\hat{Y}$ the predicted output by the decoder. The pseudo-code for our proposed approach is shown in \Algo{proposed}. 
The line 3 represents the process done by the $i$-th encoder layer, $L^{enc}_{i}$, and the line 5 does the same for the $j$-th decoder layer, $L^{dec}_{j}$.
In simple words, we compute a loss using the output of each of the $M$ decoder layers which in turn is computed using the output of each of the $N$ encoder layers. In line 10, the {\NxM} losses are aggregated\footnote{We averaged multiple losses in our experiment, but there are a number of options, such as weighted averaging.} before back-propagation. Henceforth, we will refer to this as the {\NxM} model.

For a comparison, the vanilla model is formulated in \Algo{vanilla}.

\begin{algorithm}[t]
\small
\caption{Training an {\NxM} model}
\label{algo:proposed}
\begin{algorithmic}[1]
\State $enc_{0} = X$

\For{$i$ in 1 to $N$}
    \State $enc_{i} = L^{enc}_{i}(enc_{i-1})$
    \For{$j$ in 1 to $M$}
        \State $dec_{j}=L^{dec}_{j}(dec_{j-1},enc_{i})$
        \State $\hat{Y}=softmax(dec_{j})$
        
        \State $loss_{i,j}=cross\_entropy(\hat{Y},Y)$
    \EndFor
\EndFor

\State $overall\_loss=aggregate(loss_{1,1},\ldots,loss_{N,M})$
\State Back-propagate using $overall\_loss$
\end{algorithmic}
\end{algorithm}

\begin{table*}[t]
\centering
\small

\begin{tabular}{c|cccccc|cccccc}
\hline\hline
\multirow{2}{*}{$n{\backslash}m$}
    &\multicolumn{6}{c|}{36 individual vanilla models}
    &\multicolumn{6}{c}{Our single {\NxM} model}\\
	&1	&2	&3	&4	&5	&6	&1	&2	&3	&4	&5	&6\\\hline
1 & 27.07 & 30.25 & 31.63 & 31.61 & 31.48 & 32.11 & 24.24 & 28.85 & 30.24 & 30.55 & 30.91 & 30.93\\
2 & 29.12 & 32.05 & 32.74 & 32.94 & 33.09 & 32.81 & 27.11 & 31.63 & 33.00 & 33.36 & 33.61 & 33.76\\
3 & 29.64 & 32.64 & 33.36 & 33.92 & 34.09 & 33.80 & 28.31 & 32.79 & 34.11 & 34.52 & 34.73 & 34.64\\
4 & 30.29 & 33.61 & 34.33 & 34.44 & 34.16 & 34.39 & 28.96 & 33.30 & 34.48 & 34.76 & 34.81 & 34.76\\
5 & 31.00 & 33.94 & 34.37 & 35.27 & 34.08 & 34.94 & 29.19 & 33.47 & 34.52 & 34.71 & 34.95 & 34.89\\
6 & 31.48 & 34.07 & 34.31 & 35.35 & 34.71 & 34.87 & 29.35 & 33.61 & 34.52 & 34.61 & 34.91 & 34.87\\

\hline
\end{tabular}
\caption{BLEU scores for the WMT \task{En}{De} task. The scores on the left side are for the 36 individual models that are trained separately. The scores on the right are for our proposed {\NxM} model.}
\label{table:wmt-results}

\bigskip

\begin{tabular}{c|cccccc}
\hline\hline
$n{\backslash}m$	&1	&2	&3	&4	&5	&6\\\hline
1 & 95.86 & 110.85 & 148.61 & 181.99 & 214.49 & 247.78\\
2 & 95.47 & 114.44 & 155.78 & 182.00 & 223.05 & 257.47\\
3 & 92.31 & 114.82 & 153.35 & 192.75 & 225.72 & 265.33\\
4 & 94.22 & 116.28 & 151.79 & 198.58 & 223.57 & 264.81\\
5 & 95.38 & 116.57 & 157.17 & 198.48 & 245.10 & 259.94\\
6 & 94.30 & 117.05 & 155.39 & 195.85 & 241.72 & 264.76\\

\hline
  \end{tabular}
  \caption{Decoding time (in seconds) with the different layer configurations. Given that a vanilla model with $n$ encoder and $m$ decoder layers and our {\NxM} model used with $n$ encoder and $m$ decoder layers have no difference in the amount of computation, we show only one set of decoding times.}
  \label{table:wmt-times}
\end{table*}

\section{Experiments}
\label{section:experiments}

We trained following two types of models, and evaluated them on both translation quality and decoding speed.
\begin{description}\itemsep=0mm
\item[Vanilla model:] 36 vanilla models with 1 to 6 encoder and 1 to 6 decoder layers, trained referring only to the last layer for computing loss.
\item[{\NxM} model:] A single {\NxM} model with $N=6$ encoder and $M=6$ decoder layers, trained by our multi-layer softmaxing.
\end{description}

\subsection{Datasets and Preprocessing}
We experimented with the WMT18 English-to-German (\task{En}{De}) translation task.  We used all the parallel corpora available for WMT18, except ParaCrawl corpus,\footnote{\url{http://www.statmt.org/wmt18/translation-task.html}} consisting of 5.58M sentence pairs as the training data and 2,998 sentences in newstest2018 as test data.

The English and German sentences were pre-processed using the \textit{tokenizer.perl} and \textit{lowercase.perl} scripts in Moses.\footnote{\url{http://www.statmt.org/moses}}

\begin{algorithm}[t]
\small
\caption{Training a vanilla model}
\label{algo:vanilla}
\begin{algorithmic}[1]
\State $enc_{0} = X$
\For{$i$ in 1 to $N$}
    \State $enc_{i} = L^{enc}_{i}(enc_{i-1})$
\EndFor
\For{$j$ in 1 to $M$}
    \State $dec_{j}=L^{dec}_{j}(dec_{j-1},enc_{N})$
\EndFor
\State $\hat{Y}=softmax(dec_{M})$
\State $loss=cross\_entropy(\hat{Y},Y)$
\State Back-propagate using $loss$
\end{algorithmic}
\end{algorithm}

\subsection{Model Training}

Our multi-layer softmaxing method was implemented on top of an open-source toolkit of the Transformer model \cite{NIPS2017_7181} in the version 1.6 branch of \textit{tensor2tensor}.\footnote{\url{https://github.com/tensorflow/tensor2tensor}} For training, we used the default model settings corresponding to \textit{transformer\_base\_single\_gpu} in the implementation, except what follows.
We used a shared sub-word vocabulary of 32k\footnote{We determined the sub-word vocabularies using the internal sub-word segmenter of tensor2tensor, for simplicity.} and trained the models for 300k iterations.
We trained the vanilla models on 1 GPU and our {\NxM} model on 2 GPUs with the halved batch size to ensure that both models see the same amount of training data.

We averaged the last 10 checkpoints saved every after 1k updates, and decoded the test sentences, fixing a beam size of 4 and length penalty, $\alpha$, of 0.6.\footnote{One can realize faster decoding by narrowing down the beam width.  This approach is orthogonal to ours and in this paper we do not insist which is superior to the other.} We evaluated our models using the BLEU metric \cite{Papineni:2002:BMA:1073083.1073135} implemented in tensor2tensor as \textit{t2t\_bleu}: case-sensitive and detokenized BLEU. We also report on the time (in seconds) consumed to translate the test set, which includes times for the model creation, loading the checkpoints, sub-word splitting and indexing, decoding, and sub-word de-indexing and merging, whereas times for detokenization are not taken into account.

Note that we did not use any development data for  two reasons.  First, we train all models for the same number of iterations.\footnote{In our opinion, this is a fair training method because it ensures that each model sees roughly the same number of training examples.} Second, we use checkpoint averaging before decoding, where using a development set for early stopping is not needed. We use this training and decoding approach, because it is known to give the best results for NMT using the Transformer implementation we use \cite{NIPS2017_7181}.

\subsection{Results}

\Tab{wmt-results} gives the BLEU scores and \Tab{wmt-times} gives the decoding times of the models. These summarize the cost-benefit property of our {\NxM} model in comparison with the results of the corresponding vanilla models.  When our {\NxM} model was used for decoding with the 5 encoder and 5 decoder layers, it achieved a BLEU score of 34.95 which is comparable with the BLEU score of 35.35 of the best vanilla model with 6-layer encoder and 4-layer decoder, even though the objective function for our proposed model is substantially more complex than the one for the vanilla model. Note that the vanilla models give significantly better results compared to our {\NxM} models, when using a single decoder layer. However, when the number of decoder layers are increased there is no statistically significant difference between the performance of vanilla models and our {\NxM} model; difference is less than 1.0 BLEU points in most configurations. We have essentially compressed 36 models into one.

Regarding the cost-benefit property of our {\NxM} model, two points must be noted:
\begin{itemize}\itemsep=0mm
\item BLEU score and decoding time increase only slightly, when we use more encoder layers.
\item The bulk of the decoding time is consumed by the decoder, since it works in an auto-regressive manner.  We can substantially cut down decoding time by using fewer decoder layers which does lead to sub-optimal translation quality.
\end{itemize}

Consider our {\NxM} model used with 4 encoder and 3 decoder layers which gives a BLEU of 34.48. Compared to the best vanilla model (with 6 encoder and 4 decoder layers; 35.35 BLEU), it can decode 1.3 times faster (151.79s vs. 195.85s) for the loss of 0.87 BLEU points. This loss in BLEU is statistically significant but in real-time low-latency scenarios, however, this certainly will not have a negative impact on the quality of service.\footnote{There are several researchers \citep{W15-5009,nakazawa-EtAl:2018:WAT2018} who have shown that BLEU score is not often correlated with actual translation quality judged through human evaluation.} For instance, when choosing our {\NxM} model used with 6 encoder and 2 decoder layers, we lose 1.74 BLEU points, but this might not have a massive impact on human evaluation. As such, we can choose this configuration and afford to decode almost twice as fast (117.05s vs 195.85s).

\begin{figure*}[t]
    \centering
    \begin{subfigure}[t]{0.48\textwidth}
        \includegraphics*[width=\textwidth]{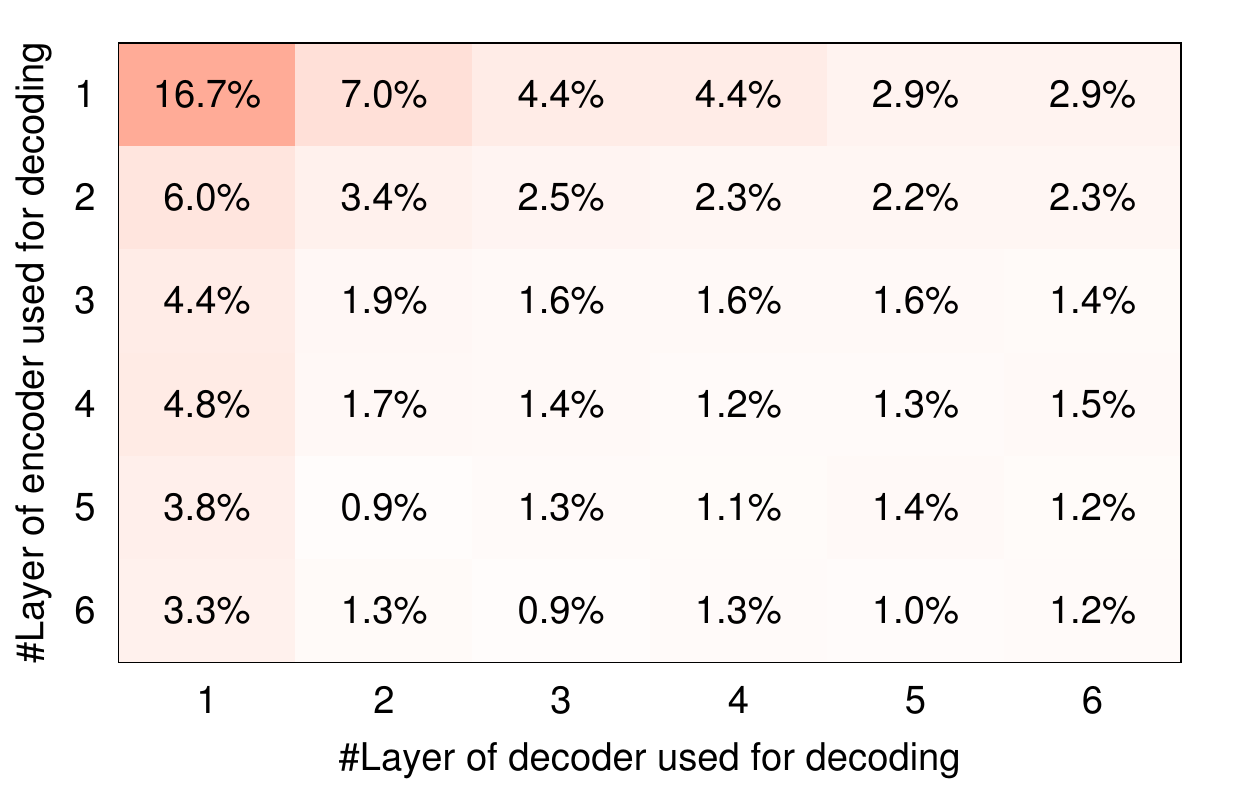}
        \subcaption{36 individual vanilla models.}
        \label{figure:wmt-vanilla-distrib-vanilla}
    \end{subfigure}
    \begin{subfigure}[t]{0.48\textwidth}
        \includegraphics*[width=\textwidth]{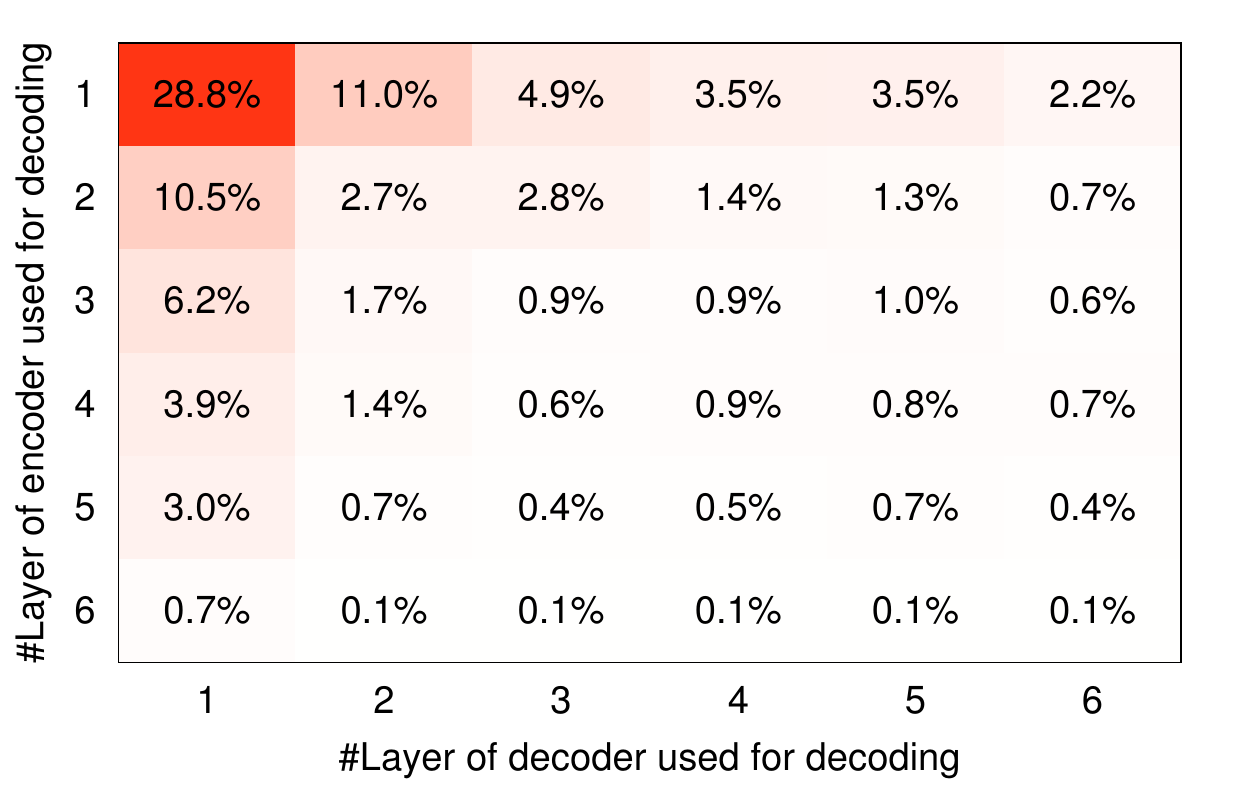}
        \subcaption{Our single {\NxM} model.}
        \label{figure:wmt-6x6-distrib}
    \end{subfigure}
    \caption{Distribution of oracle translations among 36 combinations of encoder and decoder layers for WMT \task{En}{De} (2,998 sentences).}
    \label{figure:6x6-distrib}
\end{figure*}

One may argue that training a single vanilla model with optimal numbers of encoder and decoder layers is enough. However, as discussed in \Sec{introduction}, it is impossible to know which configuration is the best a priori. More importantly, a single vanilla model cannot suffice diverse cost-benefit demands and cannot guarantee best translation for any input (see \Sec{behavior}). Recall that we aim at a flexible model and all the results in \Tab{wmt-results} have been obtained using our single {\NxM} model, albeit using different number of encoder and decoder layers for decoding.

\section{Analysis and Discussion}

In this section, to better understand the nature of our proposed method, we give an analysis of our model from the perspective of training times, model sizes, and decoding behavior, in comparison with vanilla models.

\subsection{Training Time}
All our models were trained for 300k iterations. We thus compare training times between two models by comparing the time in seconds required to complete 100 iterations of training computations.\footnote{This is the time reported by tensor2tensor by default.}  As such, the training time for our {\NxM} model was about 9.5 times that of the vanilla model with 6 encoder and 6 decoder layers. In contrast, the total training time for all individual subsumed 36 vanilla models was 25.54 times\footnote{We measured the collapsed time for a fair comparison; we assumed that all individual models are trained on a single GPU one after another, even though one may be able to use 36 GPUs to train the 36 subsumed models in parallel.} that of the vanilla model with 6 encoder and 6 decoder layers. Note that this time is calculated by adding the times required to complete 100 iterations of training computations for all individual vanilla models. Consequently, our proposed method of training an {\NxM} model is computationally much more efficient than independently training all the 36 subsumed models with different number of layers.

\subsection{Model Size}
Our proposed method can help train {\NxM} models whose number of parameters is exactly same as vanilla model with $N$ encoder and $M$ decoder layers.
If we train a set of separate models with different numbers of encoder and decoder layers, we end up with significantly more parameters.  For instance, in case of $N=M=6$ in our experiment, we have 25.16 times more parameters; a total of 4,600M for 36 subsumed models against 183M for our {\NxM} model.

\subsection{Decoding Behavior}
\label{section:behavior}

\Fig{6x6-distrib} gives the distribution of the test sentences which were best translated (oracle translations) by different combinations of encoder and decoder layers during decoding with the vanilla and our {\NxM} models.  We observed the followings.

\begin{itemize}\itemsep=0mm

\item Using the {\NxM} model, around 50\% of the test set is best translated by using 1 to 2 encoder and decoder layers despite the low corpus-level BLEU scores with these configurations.

\item In contrast, among the individual 36 models, those with 1 to 2 encoder and 1 to 2 decoder layers give the best translation for only 30\% of the test set. To cover 50\% of the test set, we have to consider the models up to 3 encoder and 3 decoder layers.
\item The distribution of best performing combinations is quite sharp for the {\NxM} model unlike the individual vanilla models.
\end{itemize}

Currently, we do not have an explanation for the difference in behavior between our and vanilla models in terms of the distribution of optimal layer combinations. However, it is clear that our {\NxM} model can essentially do what 36 individually trained models can do.

In addition, if we can predict an appropriate layer combination to decode each given input, we can automatically decode with a variable number of layers and save significant amount of computation. We leave further analyses and the design of the layer choosing mechanism for future work.

\section{Conclusion}
In this paper, we have proposed a novel procedure for training encoder-decoder models, where we softmax the output of each of the $M$ decoder layers derived using the output of each of the $N$ encoder layers. This compresses {\NxM} models into a single model that can be used for decoding with a variable number of encoder ($\le N$) and decoder ($\le M$) layers. This model can be used in different latency scenarios and hence is highly versatile. We have experimented with NMT as a case study of encoder-decoder models and given a cost-benefit analysis of our method.

In our future work, we will make an in-depth analysis on the nature of our {\NxM} models, such as the diversity of hypotheses generated by different layers. We will focus on approaches to automatically choose layer combinations depending on the input and thereby save decoding time by performing the minimal number of computations to obtain the best output.
For further speed up in decoding as well as model compaction, we plan to combine our approach with other techniques, such as those mentioned in \Sec{relwork}. Although we have only tested our idea for NMT, it should be applicable to other tasks based on deep neural networks.

\bibliography{naaclhlt2019}
\bibliographystyle{acl_natbib}

\end{document}